\pdfoutput=1

\documentclass[11pt]{article}

\usepackage{EACL2023}

\usepackage{times}
\usepackage{latexsym}

\usepackage[T1]{fontenc}

\usepackage[utf8]{inputenc}

\usepackage{microtype}

\usepackage{inconsolata}

\usepackage{booktabs}
\usepackage{multirow}
\usepackage{float}
\usepackage{graphicx}
\usepackage{amsmath}
\usepackage{url}
\usepackage{bm}
\usepackage{fdsymbol}

\renewcommand{\tt}[1]{\fontfamily{cmtt}\selectfont #1}
\title{MCoNaLa: A Benchmark for Code Generation \\ from Multiple Natural Languages}

\author{
	Zhiruo Wang\thanks{$\ $ Equal contribution.}$\ \ ^{\spadesuit}$\quad
	Grace Cuenca$^{* \vardiamondsuit}$\quad
	Shuyan Zhou$^{\spadesuit}$\quad
	Frank F. Xu$^{\spadesuit}$\quad
	{Graham Neubig}$^{\spadesuit\clubsuit}$ \\
	$^{\spadesuit}$Carnegie Mellon University \quad 
	$^{\vardiamondsuit}$Princeton University \quad 
	$^{\clubsuit}$Inspired Cognition \\
	{\tt \{zhiruow,shuyanzh,fangzhex,gneubig\}@cs.cmu.edu, gcuenca@princeton.edu}\\ 
}

\begin{document}
\maketitle
\begin{abstract}
While there has been a recent burgeoning of applications at the intersection of natural and programming languages, such as code generation and code summarization, these applications are usually English-centric. This creates a barrier for program developers who are not proficient in English. 
To mitigate this gap in technology development across languages, we propose a multilingual dataset, MCoNaLa, to benchmark code generation from natural language commands extending beyond English. Modeled off of the methodology from the English Code/Natural Language Challenge (CoNaLa) dataset, we annotated a total of 896 NL-Code pairs in three languages: Spanish, Japanese, and Russian. We present a systematic evaluation on MCoNaLa by testing state-of-the-art code generation systems. Although the difficulties vary across three languages, all systems lag significantly behind their English counterparts, revealing the challenges in adapting code generation to new languages.\footnote{Code and data are available at \url{https://github.com/zorazrw/multilingual-conala}}
\end{abstract}

\section{Introduction}
\label{sec:intro}
There are an increasing number of applications related to ``code intelligence'', such as code summarization~\citep{allamanis2016convolutional,hu2018summarizing,ahmad2020transformer} and natural language (NL) to code generation~\citep{ling2016latent,rabinovich2017abstract,yin2018mining,xu2020incorporating,norouzi2021code,wang2021codet5}, accompanied by code-specific tasks and benchmarks~\citep{oda2015learning,zhong2017seq2sql,yin2018learning,lu2021codexglue}.
However, in the cases where these benchmarks include natural language, that language is almost invariably English.

\begin{figure}
\vspace{-2mm}
    \centering
    \includegraphics[width=7.8cm]{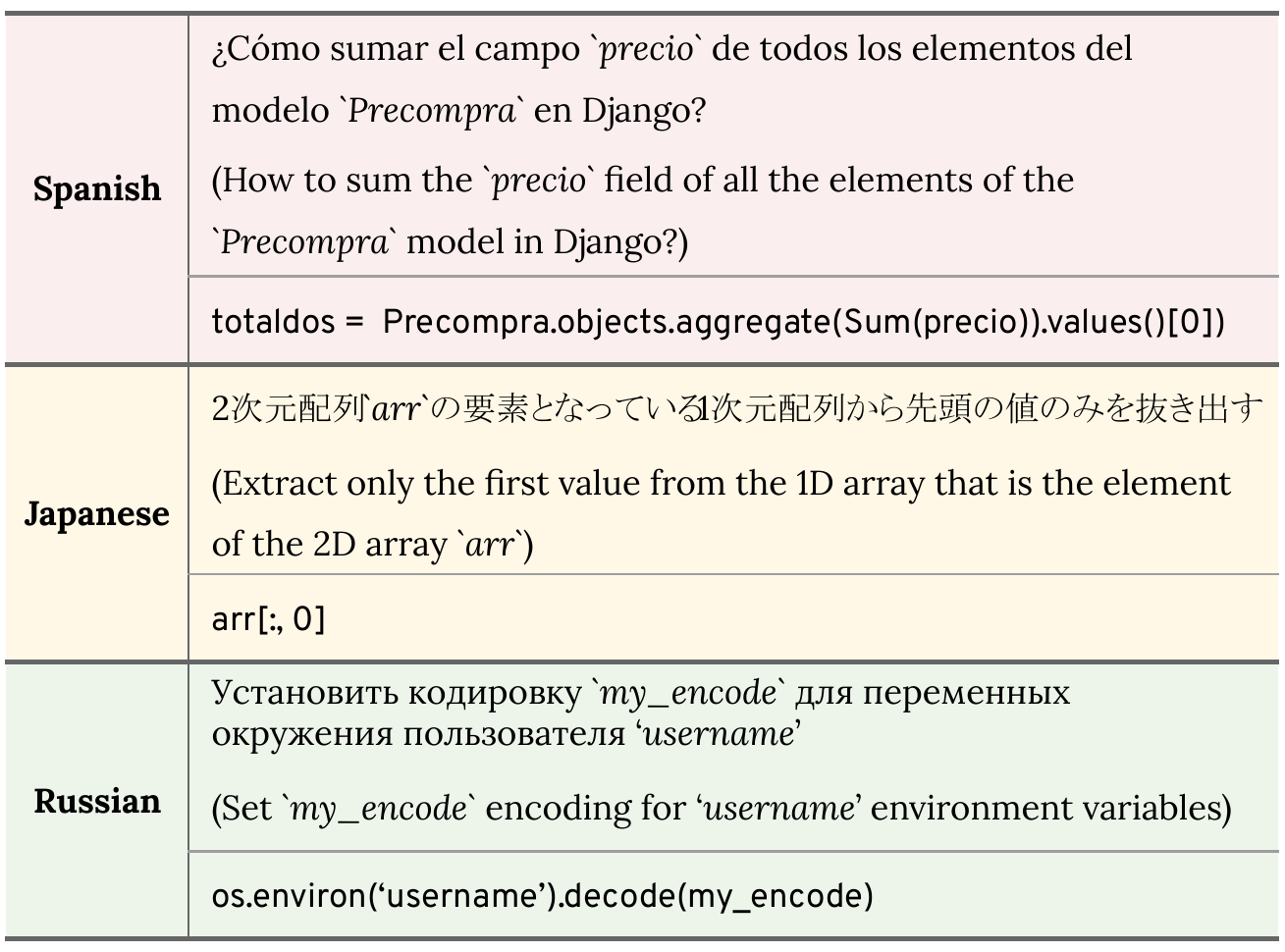}
    \caption{Examples in the MCoNaLa dataset, that aim to generate general-purpose Python code snippets from source intent of multiple natural languages. }
    \label{fig:intro-example}
\vspace{-3mm}
\end{figure}

There are a few exceptions, but most of them either focus on languages of specific domains~\citep{sherborne2021zero,sherborne2020bootstrapping,moradshahi2020localizing} or types of code~\citep{oda2015learning,liang2021lyra}, or contain NL intents collected via automatic translation \citep{li2020mtop} (\autoref{app2:related}).
However, similarly to how \citet{kwiatkowski2019natural} argue that ``natural questions'' are necessary to appropriately benchmark QA systems, we argue that ensuring the naturalness and coverage of questions is essential for benchmarking code generation systems as well.

A dataset for English code generation based on natural programming questions is the CoNaLa dataset \citep{yin2018mining}.
It is based on natural developer questions harvested from the Stack Overflow (SO) question answering forum.
In fact, in addition to English, SO also supports four other languages (Spanish, Portuguese, Japanese, and Russian) that have strong developer communities and engage in non-English programming environments.
In this work, we utilize this resource to construct the MCoNaLa dataset, consisting of $341$, $210$, and $345$ manually curated parallel samples with natural intents in Spanish, Japanese, and Russian, along with corresponding Python code snippets.
Like CoNaLa, these snippets are collected from language-specific SO sites and annotated by native speakers who are also proficient in the Python programming language. 

To provide insights in the state of code generation on this new resource, we conduct comprehensive experiments with three state-of-the-art text generation models in the context of cross-lingual transfer, by unifying training and testing NL via translation~\cite{ruder2021multi,shi2021cross,shima2010bootstrap,hartrumpf2008efficient}, or utilizing a multilingual NL encoder such as \textsc{mBART}~\citep{liu2020multilingual}. 
Our results suggest that cross-lingual NL-to-Code generation is challenging. 
Among all languages and experiment settings, the highest average BLEU score is $7.28$, far behind that of English, which achieves $33.41$, presumably because English resembles Python more than other NLs. 
In addition, we find models with task-specific modules and training outperform generic seq2seq models, yet the discrepancy between languages are consistent across all baseline models. 
In all, our corpus and experiments demonstrate the varied difficulty of the NL-to-Code generation task under different languages, emphasizing the need to develop a language-comprehensive approach to code intelligence. 

\section{The MCoNaLa Dataset}
\label{sec:dataset}

\subsection{Task Definition} 
\label{sec:dataset-definition}
Concerning the task of answering natural language questions with machine-executable programs, our focus is to build a benchmark dataset to evaluate models for their ability to encode NL \emph{intents} in multiple languages and generate code \emph{snippets}.
For each example in \autoref{fig:intro-example}, the \emph{intent} above asks how to achieve a particular goal, and the \emph{snippet} below responds with a piece of Python code.

\subsection{Annotation Workflow}
\label{sec:dataset-annotation}

Our goal is to collect \emph{intent}-\emph{snippet} parallel data in multiple natural languages. 
In this section, we outline the main workflow for data annotation: (1) language source selection, (2) valid SO post identification, and (3) parallel sample annotation. 

\begin{figure}[t]
\vspace{-3mm}
    \centering
    \includegraphics[width=7.0cm]{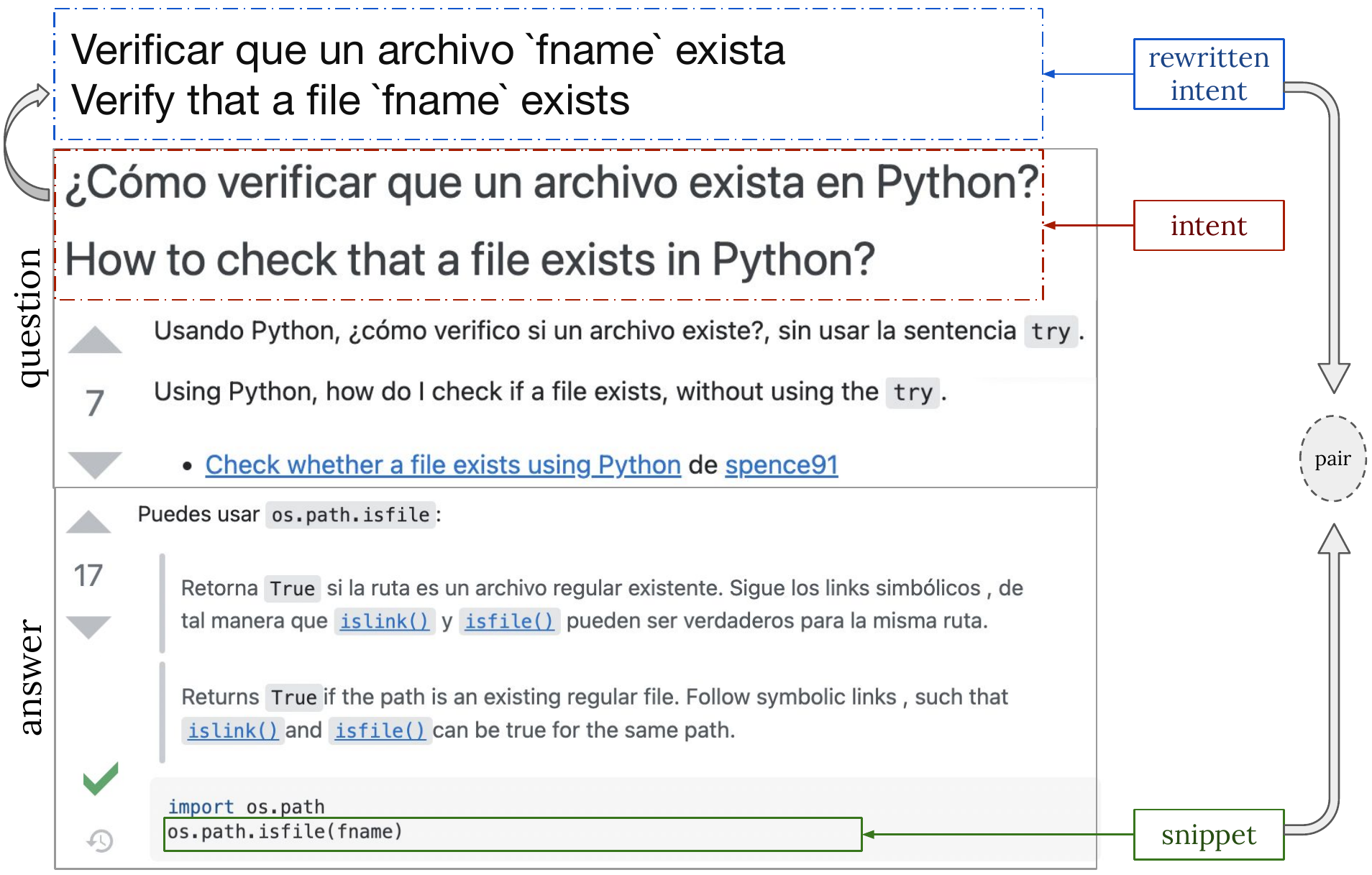}
    \caption{Illustration of the annotation process.}
    \label{fig:annotation}
\vspace{-2mm}
\end{figure}

\paragraph{Language source and selection}
Besides the English version, Stack Overflow also has forums available in four other languages: Spanish, Portuguese, Japanese, and Russian. 
Data annotation in each language requires a native speaker of that language, who should also be proficient in both English and Python. 
Due to the high cost and difficulty of hiring reliable annotators with such a specialized skill set, we only employ one Upwork annotator for each of Spanish, Japanese, and Russian. 
From the official SO data dump\footnote{\url{https://archive.org/details/stackexchange}} dated March 2021, we obtained all posts in these languages. 
However, we were unsuccessful in finding a Portuguese-speaking annotator at the time of corpus collection.

\paragraph{Identifying how-to questions} 
Following~\citet{yin2018mining}, annotators are first asked to identify valid posts that contain how-to type questions, which are imperative utterances seeking particular goals achievable by code. They are often in the post title or description, such as the example in~\autoref{fig:annotation}. 

Posts are sent in 100-sample batches, and then categorized by annotators. 
To improve annotation efficiency, we bootstrapped a \textsc{mBART} how-to question classifier, with English examples, then iteratively multilingual samples. 
It achieves an accuracy of $72.50\%$. 
We then automatically filter the probable invalid posts using this classifier and designate the rest for manual annotation. 
We collect all valid posts and extract questions as raw intents, for subsequent parallel data annotation.

\begin{figure*}
\vspace{-3mm}
    \centering
    \includegraphics[width=0.98\textwidth]{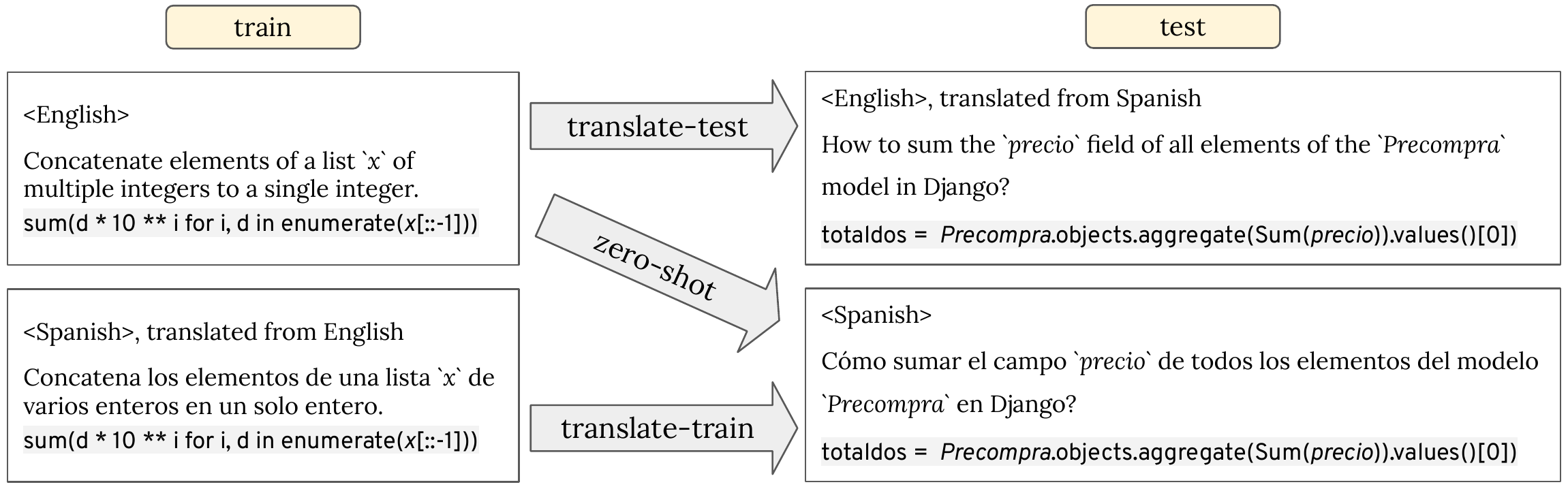}
    \caption{Example usage on the original English and Multilingual samples in three settings. }
    \label{fig:setting}
\vspace{-2mm}
\end{figure*}

\paragraph{Collecting intent-snippet pairs} 
For each post, we ask the annotators to find at most three snippets of Python code that correctly answer the extracted question. 
However, questions from post title or description are often ambiguous, especially in respective context of answer snippet, such as the example in \autoref{fig:annotation}, that the question does not specify the names of ``data'' and ``list'' variables to allow precise code implementation. 
To disambiguate the intent and align it with a snippet, we ask annotators to rewrite the intent by: (1) specifying variable names appearing in the answer snippet, and (2) clarifying commands with reference question descriptions. 
Concretely, variable names and data types in the rewritten intent need to be surrounded by the ASCII grave accent marks (e.g., \verb|`data`|), string literals or file paths should use singular typographic quotation marks (e.g., `\texttt{file1.txt}', `\texttt{https://www.abc.com/}').

The final MCoNaLa dataset consists of 341, 210, and 345 intent-snippet pairs in Spanish, Japanese, and Russian. 
It is noteworthy that the goal of MCoNaLa is to benchmark cross-lingual NL-to-Code generation task and mainly for testing purposes, instead of curating large-scale dataset for training. While its size is relatively small given the collection difficulty, we show that it can reliably inform significant method improvements (\autoref{subsec:experiment}). We believe it is important for our dataset to be representative of the naturally occurring questions in respective language environments.

\subsection{Quality Analysis}
To ensure high data quality as intended, we checked 15 random samples from each language subset. Each rater score NL intents and code snippets from 1 to 5 based on their correctness and specificity. 

The results demonstrate the high quality of our dataset, achieving 4.78, 4.65, 4.78 points on Spanish, Japanese, and Russian intents; and 4.84, 4.89, 4.78 points on their corresponding code snippets. Meanwhile, three raters present high agreement -- the Fleiss' Kappa measure is 64.29 for NL intents and 69.49 for code snippets -- both numbers indicate substantial agreement among the raters.
\section{Method}
\label{sec:method}

To provide insights about evaluating on MCoNaLa, we demonstrate potential dataset usage in three train-test settings (\autoref{subsec:method-setting}), and propose to adapt three baseline models from either multilingual (NL) or code understanding to achieve both ends (\autoref{subsec:method-model}). 

Because the size of MCoNaLa allows only testing purposes, we resort to its larger English counterpart, CoNaLa~\cite{yin2018mining}, to allow training. 
CoNaLa contains 2,879 manually annotated samples and 600$k$ samples extracted from English SO forum and API documents, which can serve as a sufficient source for training.
Given this usage, we denote the three test languages as \emph{target} languages.

\subsection{Train-Test Settings}
\label{subsec:method-setting}

We adopt three settings from two paradigms~\citep{hu2020xtreme} as illustrated in~\autoref{fig:setting}: (1) translating intents in train (\textit{translate-train}) or test (\textit{translate-test}) sets to bridge the language gap, (2) using multilingual encoder to transfer from English to target languages (\textit{zero-shot}).  

For each target language, we can align the languages of training and testing intents and use a monolingual encoder. The \textit{translate-train} setting translates English intents in CoNaLa to each target language for training and then tests with MCoNaLa samples. \textit{translate-test} translates MCoNaLa intents in three target languages into English. 
Because it is not feasible to manually translate 600$K$+ intents, we use existing multilingual machine translation (MMT) models to automate translation. We benchmarked several open-source options, as elaborated in~\autoref{sec:analysis-translation}, and settled on the \textsc{M2M-124} model used on the FLORES-101 dataset~\cite{goyal2022flores}. 

Also, we can train models on English samples and directly evaluate on MCoNaLa samples in target languages \textit{zero-shot}, requiring models to encode multiple NLs, further, transfer the code generation ability from English context to target ones. 

\subsection{Baseline Models}
\label{subsec:method-model}

We introduce three baseline methods targeting the above train-test settings. We encourage readers to refer to the original papers for more details.

In a monolingual context, models should function in target languages for \textit{translate-train} and English for \textit{translate-test}. 
\textsc{TranX}~\cite{yin2018tranx} is a BiLSTM-based encoder-decoder model that uses a transition-based abstract syntax parser to map NLs into formal meaning representations (MR) such as Python programs. 
It is agnostic to input languages and hence can be evaluated on both translated settings. 
\textsc{TAE}~\cite{norouzi2021code} is the state-of-the-art method on CoNaLa by training a generic transformer with an added target autoencoder (\textsc{TAE}) objective. 
However, it is built with (English-)BERT and is intended for English scenarios, therefore only tested on \textit{translate-test}. 

As is required by \textit{zero-shot} evaluation, we adopt a multilingual model, \textsc{mBART}~\citep{liu2020multilingual}, which is a seq2seq model pre-trained on 25 natural languages including our target ones. Note that \textsc{mBART} can also function in monolingual contexts, for both \textit{translate-train} and \textit{translate-test} settings. 

\subsection{Experiment}
\label{subsec:experiment}

We train baseline models in their available settings, then tokenize the generated and reference code snippets following~\citet{yin2018tranx} to evaluate the BLEU-4 scores. We report the average scores of five rounds using different random seeds.

\begin{table}[H]
\centering
\resizebox{0.49\textwidth}{!}{
    \begin{tabular}{ccc|cccc}
    \toprule
    \multirow{2}{*}{\textbf{Model}} & \multirow{2}{*}{\textbf{Setting}} & \multicolumn{5}{c}{\textbf{Language}} \\
    \cmidrule{3-7}
    {} & {} & \multicolumn{1}{c|}{\textbf{en}} & \textbf{es} & \textbf{ja} & \textbf{ru} & \textbf{avg.} \\ 
    \midrule
    \multirow{3}{*}{\textsc{mBART}} & {translate-test} & \multirow{3}{*}{25.20} & {2.38} & {3.07} & {2.04} & {2.50} \\
    {} & {translate-train} & {} & \textbf{2.64} & {3.45} & {2.65} & {2.91} \\
    {} & {zero-shot} & {} & {2.49} & {1.83} & {2.28} & {2.20} \\
    \midrule
    \multirow{2}{*}{\textsc{TranX}} & {translate-test} & \multirow{2}{*}{32.26} & {2.46} & {8.34} & {8.12} & {6.31} \\
    {} & {translate-train} & {} & {2.44} & {6.11} & {6.02} & {4.86} \\
    \midrule
    \textsc{TAE} & {translate-test} & \textbf{33.41} & {2.39} & \textbf{9.90} & \textbf{9.56} & \textbf{7.28}\\
    \bottomrule
    \end{tabular}
}
\caption{BLEU scores of baselines for various train-test settings in English (en) and target languages (es, ja, ru). } 
\label{tab:baseline}
\vspace{-2mm}
\end{table}

In \autoref{tab:baseline}, first, scores on target languages average to at most $7.28$, much lower than $33.41$ on English, revealing the similarity of English and Python, and the difficulty of generating code from other languages.
Second, models with code-specific designs and training (\textsc{TranX} and \textsc{TAE}) performs better in general. The lower scores of \textsc{mBART} potentially suggest a certain representation gap between NL and PL. 
Third, results on two code-specific models show consistent variations across languages: scores are lower for Spanish, but rise similarly on Japanese and Russian. As we will discuss in~\autoref{sec:analysis-lang-gap}, this is possibly due to the distributional gap between languages with varied complexity.


\subsection{Significance Test}
\label{sub:significance-test}

To verify the effectiveness of MCoNaLa, we perform significance tests~\citep{dror2018hitchhiker} to show its capability of showing significant differences between systems. We conduct paired bootstrap re-sampling tests with each pair of models in their available settings, using a sample rate of $0.5$ and a sample size of $10,000$. 

\begin{table}[H]
\centering
\resizebox{0.49\textwidth}{!}{
    \begin{tabular}{cc|ccc|c|c}
    \toprule
    \multirow{2}{*}{\textbf{Setting}} & \multirow{2}{*}{\textbf{Language}} & \multicolumn{3}{c|}{\textbf{Win Rate (\%)}} & \multirow{2}{*}{\textbf{Tie}} & \multirow{2}{*}{\textbf{p-value}} \\
    \cmidrule{3-5}
    {} & {} & \textbf{\textsc{mBART}} & \textbf{\textsc{TranX}} & \textbf{\textsc{TAE}} & {} & {} \\ 
    \midrule
    \multirow{9}{*}{translate-test} & \multirow{3}{*}{es} & {0.532} & {0.402} & {-} & {0.066} & {0.468} \\
    {} & {} & {0.522} & {-} & {0.396} & {0.102} & {0.478} \\
    {} & {} & {-} & {0.508} & {0.448} & {0.044} & {0.492} \\
    \cmidrule{2-7}
    {} & \multirow{3}{*}{ja} & {0.000} & {1.000} & {-} & {0.000} & {0.000} \\
    {} & {} & {0.000} & {-} & {1.000} & {0.000} & {0.000} \\
    {} & {} & {-} & {0.002} & {0.998} & {0.000} & {0.002} \\
    \cmidrule{2-7}
    {} & \multirow{3}{*}{ru} & {0.000} & {1.000} & {-} & {0.000} & {0.000} \\
    {} & {} & {0.000} & {-} & {1.000} & {0.000} & {0.000} \\
    {} & {} & {-} & {0.001} & {0.998} & {0.001} & {0.002} \\
    \midrule
    \multirow{3}{*}{translate-train} & {es} & {0.592} & {0.408} & {-} & {0.000} & {0.408} \\
    \cmidrule{2-7}
    {} & {ja} & {0.000} & {1.000} & {-} & {0.000} & {0.000} \\
    \cmidrule{2-7}
    {} & {ru} & {0.000} & {1.000} & {-} & {0.000} & {0.000} \\
    \bottomrule
    \end{tabular}
}
\caption{Significance testing results between each pair of baseline models. `-' marks the model not in the pair.} 
\vspace{-2mm}
\label{tab:sig-test-model}
\end{table}

In both \textit{translate-test} and \textit{translate-train} settings of \autoref{tab:sig-test-model}, code-specific systems (\textsc{TranX} and \textsc{TAE}) significantly outperform \textsc{mBART} on Japanese and Russian. However, no significant differences are shown in Spanish, as expected given its relative difficulty. 
With significance testing, one can obtain reliable results even on a small dataset. While small sizes are not entirely desirable for informative evaluation, we view them as practical reflections of data scarcity, supporting our call for more non-English resources.

\section{Analysis}
\label{sec:analysis}

\begin{figure*}[h]
\vspace{-2mm}
    \centering
    \includegraphics[width=0.92\textwidth]{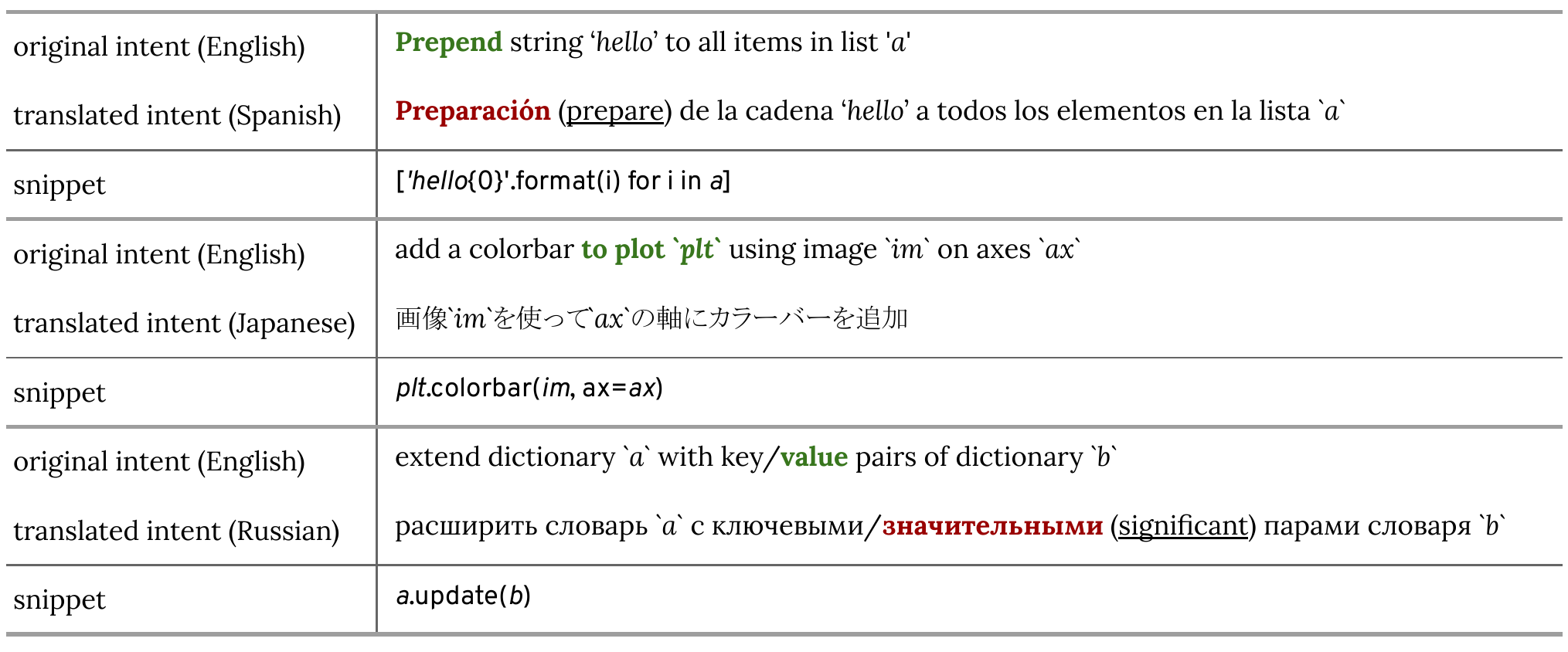}
    \caption{Examples showing that the translation errors or omits critical words in the original intent. }
    \label{fig:translation-semantic}
\end{figure*}

\subsection{Variations between Languages} 
\label{sec:analysis-lang-gap}

We first study the differences in size and snippet length between languages subsets in MCoNaLa. As listed in \autoref{tab:mconala-stats}, snippet lengths vary across languages, and the average snippet length in Spanish is around 2.5/1.3 times of that in Japanese/Russian. A longer snippet is presumably more complex, suggesting that snippets in Spanish samples are harder to generate, and hence models perform poorer.


\begin{table}[h]
\small
\centering
\resizebox{0.38\textwidth}{!}{
    \begin{tabular}{c|c|ccc}
    \toprule
    \multirow{2}{*}{\textbf{Language}} & \multirow{2}{*}{\textbf{Size}} & \multicolumn{3}{c}{\textbf{\# Snippet Tokens}} \\
    \cmidrule{3-5} 
    {} & {} & \textbf{average} & \textbf{max} & \textbf{min} \\
    \midrule
    {English} & {2,879} & {18.2} & {170} & {2} \\
    \midrule 
    {Spanish} & {341} & {42.6} & {343} & {4} \\
    {Japanese} & {210} & {17.7} & {94} & {2} \\
    {Russian} & {345} & {32.0} & {243} & {3} \\
    \bottomrule
    \end{tabular}
}
\caption{Data size and snippet length (in number of tokens) of MCoNaLa samples between target languages.}
\label{tab:mconala-stats}
\vspace{-2mm}
\end{table}

\subsection{Intent Auto-translation} 
\label{sec:analysis-translation}
In \autoref{subsec:method-setting} we use MMT models for intent translation. To optimize translation quality, we compare three best performing MMT models: OPUS-MT~\citep{TiedemannThottingal:EAMT2020}, \textsc{M2M-100}~\citep{fan2021beyond}, and \textsc{M2M-124} used in FLORES-101~\citep{goyal2022flores}. 
Since comparing in \textit{translate-train} needs intensive re-training with different model translations, we ablate in the \textit{translate-test} setting, using each model to translate test intents and evaluate NL-to-Code respectively.


\begin{table}[H]
\vspace{-1mm}
\small
\centering
\resizebox{0.46\textwidth}{!}{
    \begin{tabular}{ccccc}
    \toprule
    \multirow{2}{*}{\textbf{Baseline}} & \multirow{2}{*}{\textbf{MMT}} & \multicolumn{3}{c}{\textbf{Language}} \\
    \cmidrule{3-5}
    {} & {} & \textbf{Spanish} & \textbf{Japanese} & \textbf{Russian} \\ 
    \midrule
    \multirow{3}{*}{\textsc{mBART}} & {\textsc{M2M-124}} & {2.38} & {3.08} & {2.04} \\
    {} & {\textsc{OPUS-MT}} & {2.28} & {3.21} & {2.46} \\
    {} & {\textsc{M2M-100}} & {1.83} & {2.79} & {2.00} \\
    \midrule
    \multirow{3}{*}{\textsc{TranX}} & {\textsc{M2M-124}} & {2.46} & {8.41} & {8.09} \\
    {} & {\textsc{OPUS-MT}} & {2.46} & {5.09} & {5.00} \\
    {} & {\textsc{M2M-100}} & {2.04} & {7.38} & {8.48} \\
    \midrule
    \multirow{3}{*}{\textsc{TAE}} & {\textsc{M2M-124}} & {2.39} & {9.88} & {9.57} \\
    {} & {\textsc{OPUS-MT}} & {3.15} & {3.89} & {5.30} \\
    {} & {\textsc{M2M-100}} & {2.21} & {8.20} & {9.32} \\
    \bottomrule
    \end{tabular}
}
\caption{Comparing MMT models under \textit{translate-test}. }
\label{tab:mt-ablation}
\vspace{-1mm}
\end{table}

As in \autoref{tab:mt-ablation}, their results are close, but \textsc{M2M-124} tends to be more stable across languages and baselines.
Despite its relative superiority, its translation quality may still lag behind human performance, with more examples in \autoref{sub:translation}.

\subsection{Quality of Auto-translation}
\label{sub:translation}


To better measure the quality of translated intents, we manually check the semantic alignment between the original and translated intents, with the assistance of the Google Translate API and dictionaries. 
Concretely, we take 20 English CoNaLa intents and check if their semantics preserve after being translated into three target languages (\textit{translate-train}). We similarly examine 20 MCoNaLa intents in each target language and check their English translations (\textit{translate-test}). We use the \textsc{M2M-124} translations given its best results. 
As shown in~\autoref{fig:translation-semantic}, MMT translations are still sub-optimal: often mis-translate, even omit, the key words. This is especially severe on verbs that indicate certain Python operations. Hence, the translation step may impair intent-snippet alignment, being one of the major factors to the poor results in translated settings. 
\section{Conclusion}
\label{sec:conclude}
In this work, we extend the task of NL-to-Code generation from English-centric to multilingual scenarios. 
We establish the MCoNaLa benchmark that contains NL intent and code snippet pairs available in Spanish, Japanese, and Russian. Our benchmark serves for the multilingual code generation task, requiring models of both multilingual understanding and code synthesis. 
We conduct systematic experiments on three baseline models and show varying difficulty across languages and settings. We hope to reveal the necessity to develop, and serve as a solid test bed for language-comprehensive approaches regarding code intelligence. 

\section*{Acknowledgements}

We thank all the annotators for their hard work. This work was supported by the National Science Foundation under grant number 1815287.

\section*{Limitations}
Although the MCoNaLa dataset makes a first step to include more natural languages aside from English, it is currently limited to the languages supported by the StackOverflow forum, since SO provides the source data for the MCoNaLa creation. 
This can be mitigated by extending to more languages using programming forums in other languages that have a similar purpose to SO. 
Besides, MCoNaLa dataset only supports literal evaluation methods such as BLEU. 
Given the executable nature of Python programs, it is beneficial to support more evaluation metrics such as functional correctness, robustness, and conciseness.
\section*{Ethics Statement}
The MCoNaLa dataset is built to serve as a testbed for evaluating code generation systems from natural languages extending beyond English, given that an English-centric setting can harm universal accessibility to language technologies. 

We hire annotators who are proficient in target languages and assist them with clearly documented instructions, flexible annotation interfaces (e.g., Google Sheets), and automated methods (e.g., using a neural classifier to filter out possibly invalid cases) to optimize the annotation efficiency. We carefully check in line with our instructions and standards, to ensure the quality of both the question posts given and the annotation results back from our annotators.
We emphasize the differences between samples in different languages, because they are natural reflections of the questions that programmers asked in each specific language, similar to many works in fields such as multilingual question answering~\citep{clark-etal-2020-tydi} and named entity recognition~\citep{nothman2013learning}. We reckon that it is of paramount importance to evaluate on data that was originally produced in the target language, and results may be less reliable otherwise.

Nevertheless, with the advances in models capable of generating code from natural language inputs, we should be aware of the potentially harmful usage such as concealing malicious code~\cite{wallace2020concealed}, or generating code with security vulnerabilities~\cite{verdi2020empirical,pearce2021empirical}.

\newpage
\bibliography{anthology,custom}
\bibliographystyle{acl_natbib}

\newpage
\appendix

\section{Related Work}
\label{app2:related}

\paragraph{Natural Language to Code Generation Datasets}
There have been several benchmark datasets for NL-to-Code generation, such as Hearthstone~\citep{ling2016latent}, Django~\citep{oda2015learning}, CONCODE~\citep{iyer2018mapping}, and CoNaLa~\citep{yin2018mining}. 
Other examples include datasets for problem solving, such as HumanEval~\citep{chen2021evaluating}, MBPP~\citep{austin2021program}, and APPS~\citep{hendrycks2021measuring}. 
A number of methods have been proposed to mine intent-snippet pairs for the purpose of code search, summarization, or generation. While our work falls in the line of mining from SO~\citep{wong2013autocomment,iyer2016summarizing,yao2018staqc,yin2018learning}, other work also attempts to exploit other data sources such as API documentation~\citep{chatterjee2009sniff,movshovitz2013natural,xu2020incorporating}, code comments~\citep{wong2015clocom}, specialized sites~\citep{quirk2015language}, and developer communications~\citep{panichella2012mining}. 
One prior methodology to automatically collect large-scale parallel data is using heuristics to extract intent-snippet pairs~\citep{chatterjee2009sniff,wong2013autocomment,zagalsky2012example}, but this often results in compromised data quality~\citep{xu2020incorporating}. Our work resorts to a manual annotation strategy that often yields accurately aligned intent-snippet pairs. 

\paragraph{Multilingual Learning}
While the bulk of code-related tasks have their NL components in English, program developers native in other languages cannot enjoy the advances in code intelligence techniques, leading to the current lacunae in multilingual learning. Our work intends to mitigate this gap by facilitating NL-to-Code generation in multiple languages beyond English. 
To enable language understanding across multiple languages, a number of works propose to train language models with corpus in multiple languages~\citep{devlin2018mbert,liu2020multilingual,conneau2020unsupervised,xue2021mt5}. 
In addition to multilingual training, other data augmentation techniques commonly used in machine translation (MT), such as back-translation~\citep{edunov2018understanding}, monolingual~\citep{sennrich2016improving,siddhant2020leveraging} or generalized data augmentation~\citep{xia2019generalized}, also inspired our experiments. 
However, these techniques have rarely been utilized for NL-conditioned code generation. We present preliminary attempts in the experiments.

\end{document}